\documentclass[times,twocolumn,final, number]{elsarticle}
\usepackage[hyphens]{url}
\usepackage{hyperref} 
\usepackage{graphicx}
\usepackage{amsmath}
\usepackage{amssymb}
\usepackage{pdfpages}
\usepackage{float}
\usepackage{lipsum}
\usepackage{subcaption}
\usepackage{lscape}
\usepackage{rotating}
\usepackage{framed,multirow}
\usepackage{amssymb}
\usepackage{latexsym}
\usepackage{booktabs}
\usepackage[parfill]{parskip}

\newcommand{\rulesep}{\unskip\ \vrule\ }
\newcommand\tab[1][1cm]{\hspace*{#1}}

\usepackage{xcolor}
 \RequirePackage{geometry}
\definecolor{newcolor}{rgb}{.8,.349,.1}

\journal{}

\makeatletter
\def\ps@pprintTitle{%
 \let\@oddhead\@empty
 \let\@evenhead\@empty
 \def\@oddfoot{}%
 \let\@evenfoot\@oddfoot}
\makeatother

\begin{document}

\begin{frontmatter}

\title{Review: deep learning on 3D point clouds}

\author
{Saifullahi Aminu Bello $^{1, 2}$, Shangshu Yu$^{1}$, Cheng Wang$^{1}$, \textit{Senior Member, IEEE}\\ 
\vspace{2ex}
\normalsize{$^{1}$Fujian Key Laboratory of Sensing and Computing for Smart City, Xiamen University, China}\\
\normalsize{$^{2}$Kano University of Science and Technology,  Nigeria}\\
}

%\address{Fujian Key Laboratory of Sensing and Computing for Smart City, Xiamen University}

%\received{1 May 2013}
%\finalform{10 May 2013}
%\accepted{13 May 2013}
%\availableonline{15 May 2013}
%\communicated{S. Sarkar}

\begin{abstract}
Point cloud is point sets defined in 3D metric space. Point cloud has become one of the most significant data format for 3D representation. Its gaining increased popularity as a result of increased availability of acquisition devices, such as LiDAR, as well as increased application in areas such as robotics, autonomous driving, augmented and virtual reality. Deep learning is now the most powerful tool for data processing in computer vision, becoming the most preferred technique for tasks such as classification, segmentation, and detection. While deep learning techniques are mainly applied to data with a structured grid, point cloud, on the other hand, is unstructured. The unstructuredness of point clouds makes use of deep learning for its processing directly very challenging. Earlier approaches overcome this challenge by preprocessing the point cloud into a structured grid format at the cost of increased computational cost or lost of depth information. Recently, however, many state-of-the-arts deep learning techniques that directly operate on point cloud are being developed. This paper contains a survey of the recent state-of-the-art deep learning techniques that mainly focused on point cloud data. We first briefly discussed the major challenges faced when using deep learning directly on point cloud, we also briefly discussed earlier approaches which overcome the challenges by preprocessing the point cloud into a structured grid. We then give the review of the various state-of-the-art deep learning approaches that directly process point cloud in its unstructured form. We introduced the popular 3D point cloud benchmark datasets. And we also further discussed the application of deep learning in popular 3D vision tasks including classification, segmentation and detection.
\end{abstract}

\begin{keyword}
\texttt point cloud\sep deep learning \sep datasets \sep classification \sep segmentation \sep object detection
\end{keyword}

\end{frontmatter}

%\linenumbers

\section{Introduction}

We live in a three-dimensional world, but since the invention of the camera in 1888, visual information of the 3D world is being projected onto 2D images using cameras. 2D images, however, lose depth information and relative positions between two or more objects in the real world, which makes it less suitable for applications that require depth and positioning information such as robotics, autonomous driving, virtual reality and augmented reality among others. To capture the 3D world with depth information, early convention was to use stereo vision where 2 or more calibrated digital cameras are used to extract the 3D information. Point cloud is a data structure that is often used to represent 3D geometry, making it the immediate representation of the extracted 3D information from stereo vision cameras as well as of the depth map produced by RGB-D. Recently, 3D point cloud is booming as a result of increasing availability of sensing devices such as LiDAR and more recently, mobile phones with time of flight (tof) depth camera, which allow easy acquisition of the 3D world in 3D point cloud. 

Point cloud is simply a set of data points in a space. The point cloud of a scene is the set of 3D points sampled around the surface of the objects in the scene. In its simplest form, a 3D point cloud is represented by the XYZ coordinates of the points, however, additional features such as surface normal, RGB values can also be used. Point cloud is a very convenient format for representing 3d world and it has a range of application in different areas such as robotics, autonomous vehicles, augmented and virtual reality and other industrial purposes like manufacturing, building rendering e.t.c.

In the past few years, processing of point cloud for visual intelligence has been based on handcrafted features~\cite{spinimages1999, 3DfreeformB2004, shapeD2009, featH2008, FPFH2009,uniqueshape2010}. The review of handcrafted based feature learning techniques is conducted in ~\cite{comparison2014}. The handcrafted features do not require large training data and were seldom used as there were not enough point cloud data and deep learning was not popular. However, with increasing availability of acquisition devices, point cloud data is now readily available making use of deep learning for its processing feasible. However, the application of deep learning on point cloud is not easy due to the nature of the point cloud. In this paper, we review the challenges of point cloud for deep learning; the early approaches devised to overcome these challenges; and also the recent state-of-the-arts approaches that directly operate on point cloud, focusing more on the latter. This paper is intended to serve as guide to new researchers in the field of deep learning on point cloud as it presents the recent state-of-the-arts approaches of deep learning on point cloud. 

We organized the rest of the paper into the following: section 2 discussed the challenges of point cloud which makes the application of deep learning difficult. Section 3 reviewed the methods that overcome the challenges by converting the point cloud into a structured grid. Section 4 contains in-depth of the various deep learning methods that process point cloud directly. In section 5, we presented 3D point cloud benchmark datasets. We discussed the application of the various approaches in the 3D vision tasks in section 6. We summarize and conclude the paper in section 7.

\section{Challenges of deep learning on point clouds} \label{challenges}
\begin{figure*}[ht!]
\begin{subfigure}[t]{0.32\textwidth}
    \includegraphics[width=\textwidth]{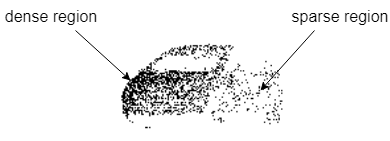}
        \caption{Irregular. Sparse and dense regions}
        \label{Irregular}
\end{subfigure}
\rulesep
\tab
\begin{subfigure}[t]{0.15\textwidth}
    \includegraphics[width=\textwidth]{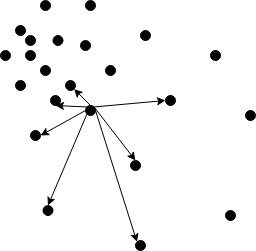}
        \caption{Unstructured. No grid, each point is independent and distance between neighboring points is not fixed}
        \label{Unstructured}
\end{subfigure}
\tab
\rulesep
\begin{subfigure}[t]{0.4\textwidth}
    \includegraphics[width=\textwidth]{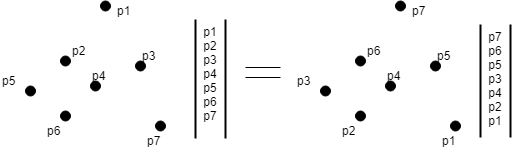}
        \caption{Unordered. As a set, point cloud is invariant to permutation}
        \label{unordered}
\end{subfigure}
\caption{Challenges}\label{fig:challenges}
\end{figure*}

Applying deep learning on 3D point cloud data comes with many challenges. Some of these challenges include occlusion which is caused by clutterd scene or blind side; noise/outliers which are unintended points; points misalignment e.t.c. However, the more pronounced challenges when it comes to application of deep learning on point clouds can be categorized into the following:

{ \bf Irregularity:} Point cloud data is also irregular, meaning, the points are not evenly sampled accross the different regions of an object/scene, so some regions could have dense points while others sparse points. These can be seen in figure ~\ref{Irregular}.

{ \bf Unstructured:} Point cloud data is not on a regular grid. Each point is scanned independently and its distance to neighboring points is not always fixed, in contrast, pixels in images are represented on a 2 dimension grid, and spacing between two adjacent pixels is always fixed.

{ \bf Unorderdness:} Point cloud of a scene is the set of points(usually represented by XYZ) obtained around the objects in the scene and are usually stored as a list in a file. As a set, the order in which the points are stored does not change the scene represented. For illustration purpose, we show the unordered nature of point sets in figure ~\ref{unordered}

These properties of point cloud are very challenging for deep learning, especially convolutional neural networks (CNN). These is because convolutional neural networks are based on convolution operation which is performed on a data that is ordered, regular and on a structured grid. Early approaches overcome these challenges by converting the point cloud into a structured grid format, section ~\ref{grid}. However, recently researchers have been developing approaches that directly uses the power of deep learning on raw point cloud, see section ~\ref{nongrid}, doing away with the need for conversion to structured grid.

\section{Structured grid based learning}  \label{grid}
\begin{figure}[H]

   \includegraphics[width=\columnwidth]{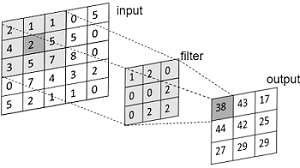}

   \caption{A typical 2D convolution operation}
\label{convolution}
\end{figure}

Deep learning, specifically convolutional neural network is successful because of the convolution operation. Convolution operation is used for feature learning, doing away with handcrafted features. Figure ~\ref{convolution} shows a typical convolution operation on a 2D grid. The convolusion operation requires a structured grid. Point cloud data on the other hand is unstructured, and this is a challenge for deep learning, and to overcome the challenge many approaches convert the point cloud data into a structured form. These approaches can be broadly divided into two categories, voxel based and multiview based. In this section, we review some of the state-of-the-arts methods in both voxel based and multiview based categories, there advantages as well as there drawbacks.

\subsection{Voxel based} ~\label{voxelb}
\begin{figure}[H]

   \includegraphics[width=\columnwidth]{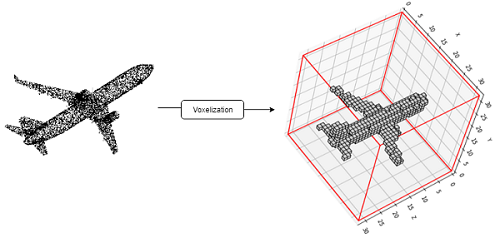}

   \caption{Point cloud of an airplane is voxelized to $30 \times 30\times 30$ volumetric occupancy grid}
\label{voxelization}
\end{figure}

Convolution operation on 2d images, uses a 2d filter of size $\dot{x} \times \dot{y}$ to convolve a 2D input represented as matrix of size $\dot{X} \times \dot{Y}$ with $\dot{x}<= \dot{X}$ and $\dot{y} <= \dot{Y}$. Voxel based methods ~\cite{3Dconvforlandingzone, voxnet, multiviewandvolumetric, normalnet, multiresolution3Dcnn} uses similar approach by converting the point cloud into a 3D voxel structure of size $X \times Y \times Z$ and convolve it with 3D kernels of size $x \times y \times z$ with $x,y,z<=X,Y,Z$ respectively. Basically, two important operations takes place in this methods, the offline(preprocessing) and the online (learning). 
The offline methods converts the point cloud into a fixed size voxels as shown in figure ~\ref{voxelization}. Binary voxels ~\cite{modelnet} is often used to represent the voxels. In ~\cite{normalnet} a normal vector is added to each voxel to improve discimination capability. 

The online operation, is the learning stage. In this stage, deep convolutional neural network is designed usually using a number of 3D convolutional, pooling, and fully connected layers.

~\cite{modelnet} represented 3D shapes as a probability distribution of binary variables on a 3D voxel grid and were the first work that uses 3D Deep Convolutional Neural Networks. The input to the network, point cloud, CAD models or RGB-D images, is converted into a 3D binay voxel grid and is processed using a convolusional deep belief network ~\cite{deepbeliefnet}. ~\cite{3Dconvforlandingzone} uses 3D CNN for landing zone detection for unmanned rotorcraft. LiDAR from the rotorcraft is used to obtain point cloud of the landing site, which is then voxelized into 3D volumes and 3D CNN  binary classifier is applied to classify the landing site as safe or otherwise. In ~\cite{voxnet} a 3D Convolutional Neural Network is proposed for object recognition, like ~\cite{modelnet}, the input to the network in ~\cite{voxnet} is converted into a 3D binary occupancy grid before applying 3D convolution operations to generate a feature vector which is passed through fully connected layers to obtain class scores. Two voxel based models where proposed in ~\cite{multiviewandvolumetric}. First model addressed overfitting using auxiliary training tasks to predict object from partial subvolumes and the second model mimic Multiview-CNNs by convolving the 3D shapes with anisotropic probing kernel.

Voxel based methods, although have shown good performances, they however do suffer from high memory consumption due to the sparsity of the voxels, figure ~\ref{voxelization}, which results in wasted computation when convolving over the non occupied regions. The memory consumption also limits the voxel resolution to usually between 32 cube to 64 cube. These drawbacks is also in addition to the artifacts introduced by the voxelization operation.

To overcome the challenges of voxelization, ~\cite{OctNet, octree}  proposed adaptive representation. These representation is much complex than the regular 3D voxels, however, its still limited to only 256 cube voxels.

\subsection{multiview based} ~\label{multiv}
\begin{figure}[H]

   \includegraphics[width=\columnwidth]{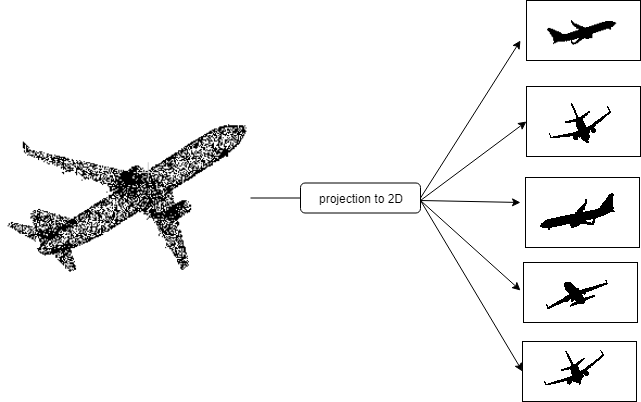}

   \caption{multiview projection of point cloud to 2D images. Each 2D image represents the same object viewed from a different angle}
\label{multiview}
\end{figure}

These methods ~\cite{multiviewCNN, SLCAE, GIFT, multiviewandvolumetric, 3Dshapeseg, classificationSphericalProjections, multiviewrecog}, take advantage of the already matured 2D CNNs into 3D. Because images are actually representation of the 3D world squashed onto a 2D grid by a camera, methods under this category follows these technique by converting point cloud data into a collection of 2D images and apply existing 2D CNN techniques to it, see ~\ref{multiview}. Compared to their volumetric based counter parts, Multiview based methods have better performance as the Multiview images contains richer information than 3D voxels even though the latter contains depth information.

~\cite{multiviewCNN} is the first work in this direction with the aim of bypassing the need for 3D descriptors for recognition and achieved state-of-the-arts accuracy. ~\cite{SLCAE} proposed a stacked local convolutional autoencoder (SLCAE) for 3D object retrieval. ~\cite{multiviewandvolumetric} introduced multi-resolution filtering which captures information at multiple scales and in addition they used data augmentation to improved on ~\cite{multiviewCNN}.

Multiview based networks have better performance than voxel based methods, this is because of two reasons, 1) they used an already well researched 2D techniques and 2) they can contains reacher information as they do not have quantization artifacts of voxelization. 

\subsection {Higher dimensional lattices}
There are other methods for point cloud processing using deep learning that converts the point cloud into higher dimensional regular lattice. SplatNet ~\cite{splatnet} processes point cloud directly, however, its primary feature learning operation occurs at the bilateral convolutional layer(BCL). The BCL layer converts the features of unordered points into a six-dimensional(6D) permutohedral lattice, and convolve it with a kernal of similar lattice. SFCNN ~\cite{sfcnn} uses a a fractalized regular icosahedral lattice to map points onto a discretized sphere and defined a multi-scale convolution operation on the regular shperical lattice.

\section{Deep learning directly on raw point cloud} \label{nongrid}
Deep learning on raw point cloud is receiving lot of attention since PointNet ~\cite{pointnet} was released in 2017. Many state-of-the-arts methods have been developed since then. These techniques process point cloud directly despite the challenges of section  ~\ref{challenges}. In this section, we review the state-of-the-arts techniques that work in this direction. We began with PointNet which is the bedrock for most of the techniques. Other techniques improved on PointNet by modeling local region structure.

\begin{figure*}

   \includegraphics[width=18cm]{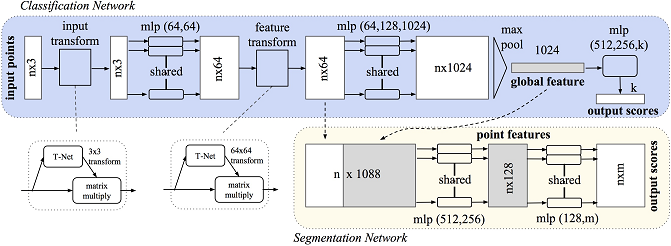}

   \caption{Architecture of PointNet ~\cite{pointnet}. PointNet is composed up of multilayer perceptrons(MLPs) which are shared point-wise and 2 spatial transformer networks(STN) of $3 \times 3$ and $64 \times 64$ dimensions which learn the canonical representation of the input set. Global feature is obtained in a winner-takes-all principal and can be used for classification and segmentation tasks.}
\label{pointnetarch}
\end{figure*}

\subsection{PointNet}
Convolutional Neural Networks is largely successful because of the convolution operation, which enables learning on local regions in a hierachical manner as the network gets deeper. Convolution however, requires structured grid which is lacking in point cloud data.  PointNet ~\cite{pointnet} is the first method that applies deep learning on unstructured point cloud and its the basis for which most other techniques are based on. In this subsection we give a review of PointNet.

The architecture of PointNet is shown in figure ~\ref{pointnetarch}. The input to PointNet is raw point cloud $ P =R^{N \times D}$, where $N$ represents the number of points in the point cloud and $D$ the dimension, usually $D=3$ representing the XYZ values of each points, however additional features can be used. Because points are unordered, PointNet is made up with symmetric funtions. Symmetric functions are functions whose output are the same irrespective of the input order. PointNet is built on 2 basic symmetric functions, multilayer perceptron(MLP) with learnable parameters, and a maxpooling function. The MLPs are feature transformations that transform the feature dimension of the points from $D=3$ to $D=1024$ dimensional space and there parameters are shared by all the points in each layer. To aggregate the global feature, maxpooling symmetric function is employed to produce one global 1024-dimensional feature vector. The feature vector represent the feature descriptor of the input which can be used for recognition and segmentation tasks.

 PointNet achieves state-of-the-arts performance on several benchmark datasets. The design of PointNet, however, do not considers local dependency among points, thus, it does not capture local structure. The global maxpooling applied select the feature vector in a "winner\textendash take \textendash all"~\cite{WTA} principle, making it very susceptible to targetted adversarial attack as demonstrated in ~\cite{pointattack}. After PointNet many approaches were proposed to capture local structure. 

\subsection{Approaches with local structure computation}
Many state-of-the-arts approaches where developed after PointNet that captures local structure. These techniques capture local structure hierarchically in a smilar fashion to grid convolution with each heirachy encoding richer representation.

\begin{figure}[H]

   \includegraphics[width=\columnwidth]{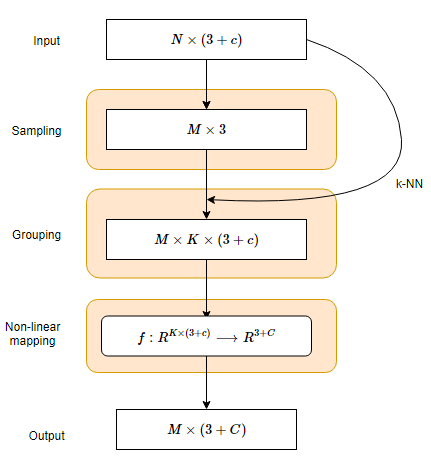}

   \caption{Basic operations for capturing local structure in point cloud. Giving $P \in R^{N \times (3+c)}$ points, each point represented by XYZ and c feature channel (for input points, c can be point features such as normals, rgb, etc or zero).  $M\leqslant N$ centroids points are sampled from $N$, and k-NN points to each of the centroid are selected to form a M groups. Each of the M group represents a local region(receptive field). A non-linear function, usually approximated by PointNet based MLP, is then applied on the local region to learn C- dimensional local region feature ($C \geqslant c$).}
\label{pointcon}
\end{figure}

Basically, due to the inherent nature of point cloud of unorderedness, local structure modeling rests on three basic operations: sampling; grouping; and a mapping function that is usually approximated by a multilayer perceptron (MLP) which maps the features of the nearest neighbor points into a feature representation that encodes higher level information, see figure~\ref{pointcon}. We briefly explained this operations before reviewing the various approaches.

\textit{Sampling} Sampling is employed to reduce resolution of points accross layers in synonymity to how convolution operation reduces the resolution of feature maps via convolutional and pooling layers. Giving point cloud $P \in R^{N \times 3}$ of N points, the sampling reduces it to M points $\hat{P} \in R^{M \times 3}$, where $M \leq N$.  The subsampled $M$ points, also referred to as representative points or centroids, are used to represent the local region from which they were sampled. Two approaches are popular for subsampling 1) random point sampling, where each of the $N$ points is equally likely to be sampled and 2) farthest point sampling (FPS) where the $M$ points are sampled such that each sampled point is the most distant point from the rest of the $M-1$ points. Other sampling methods include uniform sampling and Gumbel Subset Sampling ~\cite{selfatGSS}.

\begin{figure*}[!htb]

   \includegraphics[width=18cm]{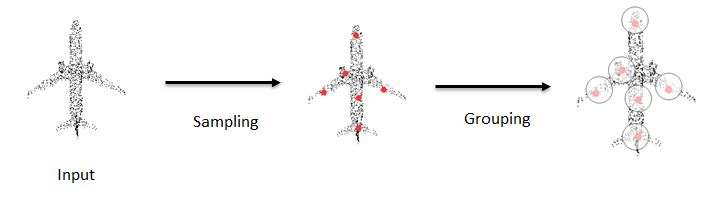}

   \caption{Sampling and Grouping of points into local patch. The reds are the centroid points selected using sampling algorithms, and the grouping shown is a ball query where points are selected based on a radius distance to the centroid.}
\label{sampgroup}
\end{figure*}

\textit{ Grouping} With the representative points sampled, k-nearest neighbor algorithm is use to select the nearest neighbor points to the representatives points to group them into a local patch, figure ~\ref{sampgroup}. The points in a local patch will be used to compute the local feature representation of the neighborhood. In grid convolution, the receptive field, are the pixels on the feature map under a kernel.  The kNN is either used directly where k nearest points to a centroid are sampled, or a ball query is used. With ball query, points are selected only when they are within a radius distance to the centroid points.

\textit{Non-linear mapping function} Once the nearest points to each representative points are obtained, the next step is to map them into a feature vector which represents the local structure. In grid convolution, the  receptive field is mapped into a feature neuron using a simple matrix multiplication and summation with convolutional kernels. This is not easy in point cloud, because the points are not structured, therefore most approaches approximate the function using PointNet ~\cite{pointnet} based methods which is composed of symmetric functions consisting of a multilayer perceptrons, $h(\cdot)$, and a maxpooling function,  $g(\cdot)$ as shown in equation~\ref{mlp}.
\begin{equation} \label{mlp} f(\{x_1, ...x_k\}) \approx g(h(x_1), . . ., h(x_k)) \end{equation}

\subsubsection{Approaches that do not explore local correlation}
Several approaches follow pointnet like MLP where correlation between points within a local region are not considered and instead, individual point features are learned via shared MLP and local region feature is aggregated using a maxpooling function in a winner-takes-all principle.

PointNet++ ~\cite{pointnetpp} extended PointNet for local region computation by applying pointnet hiearchically in local regions. Giving a point sets, $ P \in R^{N*3}$, farthest point sampling algorithm is used to select centroids, and ball query is used to select nearest neighbor points for each centroids. PointNet is then applied on the local regions to generate a feature vector of the regions. These process is repeated in a hierarchical form thereby reducing the points resolution as it goes deeper. In the last layer along the hierarchy, the whole point's features are passed through a PointNet to produce one global feature vector. PointNet++ achieves state of the art accuracy on many public datasets including, ModelNet40 ~\cite{modelnet} and ScanNet ~\cite{dai2017scannet}.

VoxelNet~\cite{voxelnet} proposed a Voxel Feature Encoding(VFE). Giving a point cloud, it is first casted into 3D voxels of resolution $\hat{D} \times \hat{H}\times \hat{W} $ , and points are grouped according to the voxel they fall into.  Because of the irregularity of point cloud, T points are sampled in each voxel inorder to have uniform number of points per voxel. In a VFE layer, the centroids for each of the voxel is computed as a local mean of the T points withing the voxel, the T points are are then processed using a fully connected network (FCN) to aggregate information from all the points similar to PointNet. The VFE layers are stacked and a maxpooling layer is applied to get a global feature vector of each voxel making the feature of the input point cloud to be represented by a sparse 4D vector, $ C \times \hat{D} \times \hat{H}\times \hat{W} $. To fit voxelnet into figure~\ref{pointcon} the centroids for each voxel are the centroids/representative points, the T points in each voxel are the nearest neighbor points and the FCN is the non linear mapping function.

Self organizing map, (SOM), originally proposed in ~\cite{som}, is used to create a self organizing networks for point cloud in SO-Net ~\cite{sonet}. While random point sampling/farthest point sampling/ uniform sampling is used to select centroids in most of the methods discussed, in So-Net, SOM is constructed with a fixed number of nodes which are dispersed uniformly in a unit ball. The SOM nodes are permutation invariant and plays the roles of local region centroid. For each SOM node, k-NN search is used to find its nearest neighbor points which are passed through a series of fully connected layers to extract point features which are maxpooled to generate M nodes features. To obtain the global feature of the input point cloud, the M nodes features are aggregated using maxpooling.

Pointwise convolution is proposed in ~\cite{pointConv}. In this technique, there is no subsampled/representative points, because the convolution operation is done on all the input points. In each point, nearest neighbor points are sampled based on a size or radius value of a kernel centered on the point. The radius value can be adjusted for different number of neighbor points in any layer. Each pointwise convolution is applied independently on the input and it transforms input points from 3-dimension to 9-dimension. The final feature is obtained by concatenating the output of all the pointwise convolution for each point and it has a resolution equavalent to the input. These final feature is then used for segmentation using convolution layer or classification task using fully connected layers.

\subsubsection{Approaches that explore local correlation}
Several approaches explore the local correlations between points in a local region to improve discriminative capability. This is intuitive because points do not exist in isolation, rather, multiple points together are needed to form a meaningful shape.

PointCNN ~\cite{pointcnn} improved on PointNet++ by proposing an X-transformation on the k-nearest neighbor points of each centroids before applying a PointNet-like MLP. The centroids/representative points are randomly sampled, and k-NN is used to select the neighborhood points which are passed through an X-transformation block before applying the non-linear mapping function. The purpose of the X-transform is to permute the input into a more canonical form which in essence also takes into consideration the relationship between points within a local region. In pointweb ~\cite{pointweb}, "a local web of points" is designed by densely connecting points within a local region and learns the impact of each point on the other points using an Adaptive Feature Adjustment (AFA) module. In ~\cite{pointconvcvpr} the authors propsed a "pointConv" operation which similarly explore the intrinsic structure of points within a local region by computing the inverse density scale of each point using a kernel density estimation (KDE). The kernel density estimation is computed offline for each point, and is fed into an MLP to estimate the density estimates. 

In ~\cite{ relationshape},  the centroids are selected using uniform sampling strategy, and the nearest neighbor points to the centroids are selected using spherical neighborhood. The non-linear function is also approximated using a multi-layer perceptron(MLP), but with additional discriminative capability by considering the relation between each centroids to its nearest neighbor points. The relationship between neighboring points is based on the spatial layout of the points. Similaryly, GeoCNN ~\cite{geocnn} explores geometric structure within local region by weighing the features of neighboring points based on the distance to their respective centroid point, however, the authors performs point wise convolution without reducing point resolution across layers.

~\cite{Acnn} argues that overlapping receptive field caused by multi-scale architecture of most of PointNet based approaches could result in computation redundancy because same neigboring points could be included in different scaled regions. To address the redundancy, the authors proposed annularly convolution which is a ring based approach that avoids having overlaps between hierarchy of receptive fields and alsp captures relationship between points in within the recpetive field.

PointNet-like MLP is the popular mapping function for approximating points in a local patch into a feature vector, however, ~\cite{spiderCNN} argues that MLP does not account for the geometric prior of point clouds and also requires sufficently large parameters. To address these issues, the authors proposed a family filters that are composed of two functions, a step function that encodes local geodesic information, followed by a third order taylor expansion. The approach learns hierarchical representations and achieves state-of-the-art performance in classification and segmentation tasks.

Point Attention transformers (PAT) is proposed in ~\cite{selfatGSS}. The authors proposed a new subsampling method termed "Gumbel Subset Sampling (GSS)" which unlike farthest point sampling (FPS), its permutation invariant, and its robust to outliers. The authors used absolute and relative position embedding, where each point is represented by a set of its absolute position and relative position to other points in a local patch, pointNet is then applied on the set. And to further capture relationship between points, a modified Multi-Head Attention (MHA) mechanism is used. A new sampling ang grouping techniques with learnable parameters were proposed in ~\cite{dpam} in a module termed dynamic points agglomeration module(DPAM) which learns an agglomeration matrix which when multiplied with incoming poimt features reduces the resolution(similar to sampling) and produce an aggregated feature (similar to grouping and pooling).

\begin{table*}
\begin{center}
\renewcommand{\arraystretch}{1.25}
\begin{tabular}[width=.3]{p{3cm}|p{5cm}|p{2.4cm}|p{6cm}}
\hline

\textbf{Method}&\textbf{Sampling} & \textbf{Grouping} & \textbf{Mapping Function} \\
\hline\hline

PointNet~\cite{pointnet} & -&-&MLP\\
\hline

PointNet++ ~\cite{pointnetpp}& Uniform subsampling&Radius-search&MLP\\
\hline
PointCNN~\cite{pointcnn} & Uniform/Random sampling&k-NN&MLP\\
\hline
So-Net~\cite{sonet} & SOM-Nodes&Radius-search&MLP\\
\hline
Pointwise Conv~\cite{pointConv} &-&Radius-search&MLP\\
\hline
Kd-Network~\cite{DeepKdNet} & - &Tree based nodes&Affine transformations+ReLU\\
\hline
DGCNN~\cite{ DGCNN} & -&k-NN&MLP\\
\hline
LocalSpec~\cite{localspecGCNN} &Farthest point sampling &k-NN&Spectral convolution+cluster pooling\\
\hline
SpiderCNN~\cite{spiderCNN} & Uniform sampling&k-NN&Taylor expansion\\
\hline
R-S CNN~\cite{relationshape} & Uniform sampling&radius-nn&MLP\\
\hline
PointConv~\cite{pointconvcvpr} & Uniform sampling&radius-nn&MLP\\
\hline
PAT~\cite{selfatGSS} & Gumbel subset sampling&k-NN&MLP\\
\hline
A-CNN~\cite{Acnn} &Uniform subsampling&k-NN&MLP+density functions\\
\hline
ShellNet~\cite{shellnet} &Random Sampling&Spherical Shells&1D convolution\\

\hline
\end{tabular}
\end{center}
\caption{Summary of methods showing sampling, grouping and the mapping function used}
\label{localregionsummary}
\end{table*}

\subsubsection{Graph based}
Graph based approaches were proposed in ~\cite{DeepKdNet, DGCNN, localspecGCNN, Point2Node}. Graph based approaches represents the point cloud with graph structure by treating each point as a node.  Graph structure is good for modelling correlation between points as explicitly represented by the graph edges. ~\cite{DeepKdNet} uses a kd-tree which is a special kind of graph. The kd-tree is built in a top-down manner on the point cloud to create a feed-forward Kd-network with learnable parameters in each layer. The computation performed in the Kd-network is in buttom-up fashion. The leaves represents the input points, 2 nearest neighbor (left and right) nodes are used to compute their parent node using shared parameters of weight matrix and a bias. The Kd-network captures hierarchical representations along the depth of the kd-tree, however,  because of tree design, nodes at the same depth level do not capture overlapping receptive fields.

~\cite{DGCNN, localspecGCNN, Point2Node} are based on typical graph network $G=\{V, E\}$ whose vertices $V$ represents the points and edges $E$ represented as a $V \times V$ matrix. In ~\cite{DGCNN} edge convolution is proposed. The graph is represented as a k-nearest neighbor graph over the inputs. In each edge convolution layer, features of each point/vertex are computed by applying a non-linear function on its nearest neighbor vertices as captured by the edge matrix $E$. The non-linear function is a multilayer perceptron (MLP). After the last edgeConv layer, global maxpooling is employed to obtain a global feature vector similar to ~\cite{pointnet}. One distinct difference of ~\cite{DGCNN} from normal graph network is that the edges are updated after each edgeConv layer based on the computed features from the previous layer hence the name Dynamic Graph CNN(DGCNN). While there is no resolution reduction as the network goes deeper in DGCNN which leads to increased in computation cost, ~\cite{localspecGCNN} defined a spectral graph convolution in which the resolution of the points reduces as the network gets deeper. In each layer, k-nearest neighbor points are sampled, but instead of using mlp-like operation on the the k local points sets like in ~\cite{pointnetpp}, a graph $G_k=\{V, E\}$ is defined on the sets, the vertices $V$ of the graph are the points and the edges $E\subseteq V\times V$ are weight based on the  pair-wise distance between the xyz spatial corrdinates of the points. Graph fourier transform of the points is then computed and filtered using spectral filtering. After the filtering, the resolution of the points is still the same, clustering, recursive cluster pooling technique is proposed to aggregate the information in each graph into one vertex.

In ~\cite{Point2Node}, the authors proposed a graph network that fully explore not only the local correlation, but also non local correlation. The correlation  is explored in 3 ways, self correlation which explores  channel-wise correlation of a node's feature; local correlation that explore local dependency among nodes in a local region; and non-local correlation for capturing better global feature by considering long-range local features.

Table ~\ref{localregionsummary} summarized the approaches showing there sampling, grouping and mapping function methods.

\section{Benchmark Datasets}\label{Benchmark}
A considerable amount of point cloud datasets has been published in recent years.
Most of the existing datasets are provided by universities and industries.
They can provide a fair comparison for testing diverse approaches.
These public benchmark datasets consist of virtual scenes or real scenes, which focus particularly on point cloud classification, segmentation, registration and object detection.
They are notably useful in deep learning since they can provide huge amounts of ground truth labels for training the network.
The point cloud is obtained by different platforms/methods, such as Structure from Motion (SfM), Red Green Blue -Depth (RGB-D) cameras, and Light Detection And Ranging (LiDAR) systems.
The availability of benchmark datasets usually decrease as the size and complexity increases.
In this section, we introduce some popular datasets for 3D research.

\subsection{3D Model Datasets}
\paragraph{ModelNet~\cite{modelnet}:}
This dataset was developed by the Princeton Vision \& Robotics Labs.
ModelNet40 has 40 man-made object categories (such as airplane, bookshelf and chair) for shape classification and recognition.
It consists of 12,311 CAD models, which has been split into 9,843 training and 2,468 testing shapes.
ModelNet10 dataset is a subset of ModelNet40 that only contains 10 categories of classes.
It is also divided into 3991 training and 908 testing shapes.

\vspace{-0.4cm}
\paragraph{ShapeNet~\cite{shapenet}:}
The large-scale dataset was developed by Stanford University et al.
It provides semantic category labels for per model.
rigid alignments, parts and bilateral symmetry planes,
physical sizes, keywords, as well as other planned annotations.
ShapeNet has indexed almost 3,000,000 models when the dataset published, and there are 220,000 models has been classified into 3,135 categories.
ShapeNetCore is a subset of ShapeNet, which consists of nearly 51,300 unique 3D models. 
It provides 55 common object categories and annotations.
ShapeNetSem is also a subset of ShapeNet, which contains 12,000 models.
It is more smaller but covers more extensive categories of 270.

\vspace{-0.4cm}
\paragraph{Augmenting ShapeNet:}
~\cite{yi2016scalable} has created detailed part labels for 31963 models from ShapeNetCore dataset.
It provides 16 shape categories for part segmentation.  
~\cite{dai2017complete} has provided 1200 virtual partial models from ShapeNet dataset.
~\cite{photoshape2018} has proposed an approach for automatically generating photorealistic materials for 3D shapes.
It is built on the ShapeNetCore dataset.
~\cite{Mo_2019_CVPR} is a large-scale dataset with fine-grained and hierarchical part annotations.
It consists of 24 object categories and 26,671 3D models, which provides 573,585 part instance labels.
~\cite{xiang2016objectnet3d} has contributed a large-scale dataset for 3D object recognition.
There are 100 categories of the dataset, which consists of 90,127 images with 201,888 objects (from ImageNet~\cite{deng2009imagenet}) and 44,147 3D shapes (from ShapeNet).

\vspace{-0.4cm}
\paragraph{Shape2Motion~\cite{wang2019shape2motion}:}
Shape2Motion was developed by Beihang University and National University of Defense Technology.
It has created a new benchmark dataset for 3D shape mobility analysis.
The benchmark consists of 45 shape categories with 2440 models where the shapes are obtained from ShapeNet and 3D Warehouse ~\cite{3DWarehouse}.
The proposed approach inputs a single 3D shape, then predicts motion part segmentation results and motion corresponding attributes jointly.

\vspace{-0.4cm}
\paragraph{ScanObjectNN~\cite{uy-scanobjectnn-iccv19}:}
ScanObjectNN was developed by Hong Kong University of Science and Technology et al.
It is the first real-world dataset for point cloud classification.
About 15,000 objects are selected from indoor datasets (SceneNN~\cite{scenenn-3dv16} and ScanNet~\cite{dai2017scannet}).
And the objects are split into 15 categories where there are 2902 unique object instances.  

\subsection{3D Indoor Datasets}
\paragraph{NYUDv2~\cite{silberman2012indoor}}
The New York University Depth Dataset v2 (NYUDv2) was developed by the New York University et al. 
The dataset provided 1449 RGB-D (obtained by Kinect v1) images captured from 464 various indoor scenes.
All of the images are distributed segmentation labels.  
This dataset is mainly served for understanding how 3D cues can lead to better segmentation for indoor objects.

\vspace{-0.4cm}
\paragraph{SUN3D~\cite{xiao2013sun3d}}
This dataset was developed by the Princeton University.
It is a RGB-D video dataset where the videos are captured from 254 different spaces in 41 buildings.
SUN3D provides 415 sequences with camera pose and object labels.
The point cloud is generated by structure from motion (SfM). 

\vspace{-0.4cm}
\paragraph{S3DIS~\cite{s3dis}}
Stanford 3D Large-Scale Indoor Spaces (S3DIS) was developed by the Stanford University et al.
S3DIS was collected from 3 different buildings with 271 rooms where the cover area is above 6,000$m^2$.
It contains over 215 million points, and each point has the provision of instance-level semantic segmentation labels (13 categories).

\vspace{-0.4cm}
\paragraph{SceneNN~\cite{scenenn-3dv16}}
Singapore University of Technology and Design et al. developed this dataset.
SceneNN is an RGB-D (obtained by Kinect v2) scene dataset collected form 101 indoor scenes. 

It provides 40 semantic classes for the indoor scenes, and all semantic labels are same as NYUDv2 dataset.

\vspace{-0.4cm}
\paragraph{ScanNet~\cite{dai2017scannet}}
ScanNet is a large-scal indoor dataset developed by Stanford University et al.
It contains 1513 scanned scenes, including nearly 2.5M RGB-D (obtained by Occipital Structure Sensor) images from 707 different indoor environments.
The dataset provides ground truth labels for 3D object classification with 17 categories and semantic segmentation with 20 categories.

For object classification, ScanNet divides all instances into 9,677 instances for training and 2,606 instances for testing.
And ScanNet splits all scans into 1201 training scenes and 312 testing scenes for semantic segmentation.

\vspace{-0.4cm}
\paragraph{Matterport3D~\cite{Matterport3D}}
Matterport3D is the largest indoor dataset which developed by Princeton University et al.
The cover area of this dataset is 219,399m$m^2$ from 2056 rooms, and there is 46,561m$m^2$ of floor space.
It consists of 10,800 panoramic views where the views are from 194,400 RGB-D images of 90 large-scale buildings.
The labels contain surface reconstructions, camera poses, and semantic segmentation. 
This dataset investigates 5 tasks for scene understanding, which are keypoint matching, view overlap prediction, surface normal estimation, region-type classification, and semantic segmentation.

\vspace{-0.4cm}
\paragraph{3DMatch~\cite{zeng20163dmatch}}
This benchmark dataset is developed by Princeton University et al.
It is a large collection of existing datasets, such as Analysisby-Synthesis~\cite{valentin2016learning}, 7-Scenes~\cite{shotton2013scene}, SUN3D~\cite{xiao2013sun3d}, RGB-D Scenes v.2~\cite{de2013unsupervised} and Halber et al.~\cite{Halber2016StructuredGR}.
3DMatch benchmark consists of 62 scenes with 54 training scenes and 8 testing scenes.
It leverages correspondence labels from RGB-D scene reconstruction datasets, and then provides ground truth labels for point cloud registration.

\vspace{-0.4cm}
\paragraph{Multisensor Indoor Mapping and Positioning Dataset~\cite{wang2018semantic}}
This indoor dataset (rooms, corridor and indoor parking lots) was developed by Xiamen University et al. 
The data was acquired by multi-sensors, such as laser scanner, camera, WIFI, Bluetooth, and IMU.
This dataset provides dense laser scanning point cloud for indoor mapping and positioning.
Meanwhile, they also provide colored laser scans based on multi-sensor calibration and SLAM-mapping process.

\subsection{3D Outdoor Datasets}
\paragraph{KITTI~\cite{Geiger2012CVPR}~\cite{Geiger2013IJRR}}
The KITTI dataset is one of the best known in the field of autonomous driving which was developed by Karlsruhe Institute of Technology et al.
It can be used for the research of stereo image, optical flow estimation, 3d detection, 3d tracking, visual odometry and so on.
The data acquisition platform is equipped with two color cameras, two grayscale cameras, a Velodyne HDL-64E 3D laser scanner and a high-precision GPS/IMU system.
KITTI provides raw data with five categories of Road, City, Residential, Campus and Person.
Depth completion and prediction benchmark consists of more than 93 thousand depth maps.
3D object detection benchmark contains 7481 training point clouds and 7518 testing point clouds.
Visual odometry benchmark is formed by 22 sequences, with 11 sequences (00-10) LiDAR data for training and 11 sequences (11-21) LiDAR data for testing. 
Meanwhile, a semantic labeling~\cite{behley2019iccv} for Kitti odometry dataset is published recently.
SemanticKITTI contains 28 classes including ground, structure, vehicle, nature, human, object, and others.

\vspace{-0.4cm}
\paragraph{ASL Dataset~\cite{pomerleau2012challenging}}
This group of datasets was developed by ETH Zurich.
The dataset was collected between August 2011 to January 2012.
It provides 8 point cloud sequences acquired by a Hokuyo UTM-30LX. 
Each sequences has around 35 scanning point clouds and the ground truth pose is supported by GPS/INS systems.
This dataset covers the area of structured and unstructured environments.  

\vspace{-0.4cm}
\paragraph{iQmulus~\cite{bredif2014terramobilita}}
The large-scale urban scene dataset was developed by Mines ParisTech et al in January 2013.
The entire 3D point cloud has been classified and segmented into 50 classes.
The data was collected by StereopolisII MLS, a system developed by French National Mapping Agency (IGN).
They use Riegl LMS-Q120i sensor to acquire 300 million points.

\vspace{-0.4cm}
\paragraph{Oxford Robotcar~\cite{RobotCarDatasetIJRR}}
This dataset was developed by the University of Oxford.
It consists of around 100 times trajectories (a total of 101,046km trajectories) through central Oxford between May 2014 to December 2015.
This long-term dataset captures many challenging environment changes including season, weather, traffic, and so on.
And the dataset provides both images, LiDAR point cloud, GPS and INS ground truth for autonomous vehicles.
The LIDRA data were collected by two SICK LMS-151 2D LiDAR scanners and one SICK LD-MRS 3D LIDAR scanner.

\vspace{-0.4cm}
\paragraph{NCLT~\cite{carlevaris2016university}}
It was developed by the University of Michigan.
It contains 27 times trajectories through the University of Michigan’s North Campus between January 2012 to April 2013.
This dataset also provides images, LiDAR, GPS and INS ground truth for long-term autonomous vehicles.
The LiDRA point cloud was collected by a Velodyne-32 LiDAR scanner.

\vspace{-0.4cm}
\paragraph{Semantic3D~\cite{hackel2017semantic3d}}
The high quality and density dataset was developed by ETH Zurich.
It contains more than four billion of points where the point cloud are acquired by static terrestrial laser scanners.
There are 8 semantic classes provided, which consist of man-made terrain, natural terrain, high vegetation, low vegetation, buildings, hard scape, scanning artefacts and cars.
And the dataset is split into 15 training scenes and 15 testing scenes.

\vspace{-0.4cm}
\paragraph{DBNet~\cite{chen2018lidar}}
This real-world LiDAR-Video dataset was developed by Xiamen University et al.
It aims at learning driving policy, since it is different from previous outdoor datasets. 
DBNet provides LiDAR point cloud, video record, GPS and driver behaviors for driving behavior study.
It contains 1,000 km driving data captured by a Velodyne laser.

\vspace{-0.4cm}
\paragraph{NPM3D~\cite{roynard2017parisIJRR}}
The Nuage de Points et Modélisation 3D (NPM3D) dataset was developed by PSL Research University.
It is a benchmark for point cloud classification and segmentation, and all point cloud has been labeled to 50 different classes. 
It contains 1,431 M points data collected in Paris and Lille.
The data was acquired by a Mobile Laser System including a Velodyne HDL-32E LiDAR and GPS/INS systems.

\vspace{-0.4cm}
\paragraph{Apollo~\cite{song2019apollocar3d}~\cite{lu2019l3}}
The Apollo was developed by Baidu Research et al and it is a large-scale autonomous driving dataset.
It provides labeled data of 3D car instance understanding, LiDAR point cloud object detection and tracking, and LiDAR-based localization.
For 3D car instance understanding task, there are 5,277 images with more than 60K car instances.
Each car has an industry-grade CAD model.
3D object detection and tracking benchmark dataset contains 53 minutes sequences for training and 50 minutes sequences for testing.
It is acquired at the frame rate of 10fps/sec and labeled at the frame rate of 2fps/sec.
The Apollo-SouthBay dataset provides LiDAR frames data for localization.
It was collected in southern San Francisco Bay Area.
They equip a high-end autonomous driving sensor suite (Velodyne HDL-64E, NovAtel ProPak6, and IMU-ISA-100C) on a standard Lincoln MKZ sedan.

\vspace{-0.4cm}
\paragraph{nuScenes~\cite{nuscenes2019}}
The nuTonomy scenes (nuScenes) dataset proposes a novel metric for 3D object detection which was developed by nuTonomy (an APTIV company).
The metric consists of multi-aspects, which are classification, velocity, size, localization, orientation, and attribute estimation of the object.
This dataset was acquired by an autonomous vehicle sensor suite (6 cameras, 5 radars and 1 lidar) with 360 degree field of view.
It contains 1000 driving scenes collected from Boston and Singapore, where the two cities are both traffic-clogged.
The objects in this dataset have 23 classes and 8 attributes, and they are all labeled with 3D bounding boxes. 

\vspace{-0.4cm}
\paragraph{BLVD~\cite{blvdICRA2019}}
This dataset was developed by Xian Jiaotong University and it was collected in Changshu (China).
It introduces a new benchmark which focuses on dynamic 4D object tracking, 5D interactive event recognition and 5D intention prediction.
BLVD dataset consists of 654 video clips, where the videos are 120k frames and the frame rate is 10fps/sec.
All frames are annotated to obtain 249,129 3D annotations.
There are totally 4,902 unique objects for tracking, 6,004 fragments for interactive event recognition, and 4,900 objects for intention prediction.

\begin{table*}[h]
  \centering

    \begin{tabular}{|c|p{3cm}|c|p{4cm}|p{4cm}|}
    \toprule
    \multicolumn{2}{|c|}{} & Model & \multicolumn{1}{c|}{Indoor} & \multicolumn{1}{c|}{Outdoor} \\
    \midrule
    \multicolumn{2}{|c|}{CAD} & \multicolumn{1}{p{3cm}|}{ModelNet (2015, cls),\newline{} ShapeNet (2015, seg), \newline{}Augmenting ShapeNet, \newline{}Shape2Motion (2019, seg, mot)} & \multicolumn{1}{c|}{} & \multicolumn{1}{c|}{} \\
    \midrule
    \multicolumn{2}{|c|}{RGB-D} & ScanObjectNN (2019, cls) & NYUDv2 (2012, seg),\newline{}SUN3D (2013, seg), \newline{}S3DIS (2016, seg), \newline{}SceneNN (2016, seg), \newline{}ScanNet (2017, seg), \newline{}Matterport3D (2017, seg), \newline{}3DMatch (2017, reg) & \multicolumn{1}{c|}{} \\
    \midrule
    \multirow{2}[4]{*}{LiDAR} & \multicolumn{1}{c|}{terrestrial LiDAR scanning} &       & \multicolumn{1}{c|}{} & \multicolumn{1}{c|}{Semantic3D (2017, seg)} \\
\cmidrule{2-5}          & mobile LiDAR scanning &       & Multisensor Indoor Mapping and Positioning Dataset (2018, loc) & KITTI (2012, det, odo),\newline{} Semantic KITTI (2019, seg),\newline{} ASL Dataset (2012, reg),\newline{} iQmulus (2014, seg),\newline{} Oxford Robotcar (2017, aut),\newline{}NCLT (2016, aut),\newline{}DBNet (2018, dri),\newline{}NPM3D (2017, seg),\newline{}Apollo (2018, det, loc),\newline{}nuScenes (2019, det, aut),\newline{}BLVD (2019, det) \\
    \bottomrule
    \end{tabular}%
  \caption{categorization of benchmark datasets. (cls: classification, 	
seg: segmentation,	
loc: localization,	
reg: registration,	
aut: autonomous driving,	
det: object detection,	
dri: driving behavior,	
mot: motion estimation,	
odo: odometry,	
)}
  \label{benchmarkdatasets}%
\end{table*}%

\section{Application of deep learning in 3D vision tasks}
\begin{figure*}[ht!]
\begin{subfigure}[t]{0.32\textwidth}
    \includegraphics[width=\textwidth]{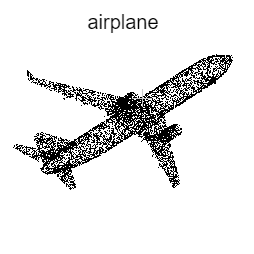}
        \caption{Object classification}
        \label{objclassificationl}
\end{subfigure}
\rulesep
\begin{subfigure}[t]{0.32\textwidth}
    \includegraphics[width=\textwidth]{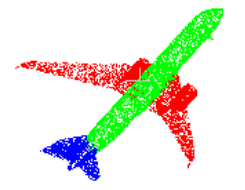}
        \caption{Parts segmentation}
        \label{partseg}
\end{subfigure}
\rulesep
\begin{subfigure}[t]{0.32\textwidth}
    \includegraphics[width=\textwidth]{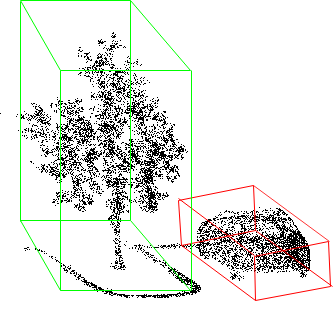}
        \caption{Object detection}
        \label{objdetectionl}
\end{subfigure}
\centering
\begin{subfigure}[t]{0.9\textwidth}
    \includegraphics[width=17cm, height=9cm]{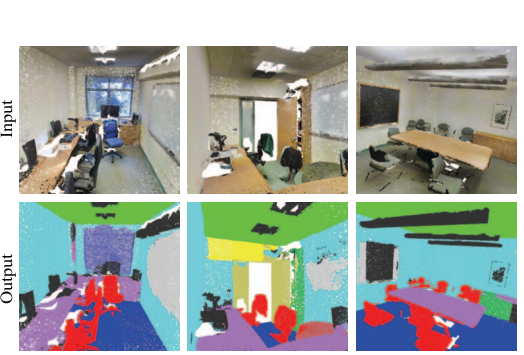}
        \caption{Semantic segmentation ~\cite{pointnet}}
        \label{insseg}
\end{subfigure}
\caption{Deep learning tasks on point cloud. (a) Object classification (b) Parts segmentation (c) Object detection (d) Semantic segmentation (Best viewed with color)} \label{fig:tasks}
\end{figure*}

In this section we discussed the application of the methods discussed in section ~\ref{nongrid} in 3 popular 3D vision tasks namely: classification, segmentation and object detection. See figure ~\ref{fig:tasks}. We review the performance of the methods on popular benchmark datasets, Modelnet40 dataset~\cite{modelnet} for classification, ShapeNet~\cite{shapenet} and Stanford 3D Indoor Semantics Dataset(S3DIS) ~\cite{s3dis} datasets for parts and semantic segmentation respectively.

\subsection{Classification}
Object classification has been one of the primary areas for which deep learning is used. In object classification the objective is: giving a point cloud, a network should classify it into a certain category. Classification is the pioneering task in deep learning because early breakthrough deep learning models such as AlexNet~\cite{alexnet}, VGGNet~\cite{VGG}, and ResNet~\cite{resnet} are classification models. In point cloud, most early techniques for classification using deep learning relied on a structured grid, section~\ref{grid}, however, we limit ourself to only approaches that process point cloud directly.

\begin{figure*}[h]

   \includegraphics[width=\textwidth, height = 9cm]{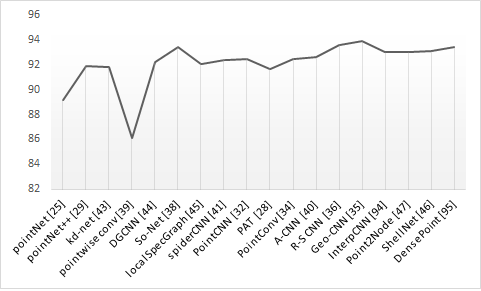}

   \caption{Timeline of classification accuracy of ModelNet40}
\label{ctimeline}
\end{figure*}

The features learned by the techniques reviewed in both section ~\ref{nongrid} and ~\ref{grid} can easily be used for classification task by passing them through a fully connected network whose last layer represents classes. Other machine learning classifiers such as SVM can also be used as in ~\cite{voxnet, foldingnet}. In figure ~\ref{ctimeline} a timeline performance of point based deep learning approaches on modelnet40 is shown.

\subsection{segmentation}
Segmentation of point cloud is the grouping of points into homegenous regions. Traditionally, segmentation is done using edges~\cite{edgeseg2006} or surface properties such as normals, curvature and orientation ~\cite{edgeseg2006, regiongrowing}. Recently, feature based deep learning approaches are used for point cloud segmentation with the goal of segmenting the points into different aspects. The aspects could be different parts of an object, referred to as part segmentation or different class categories, also referred to as semantic segmentation. 

In parts segmentation, the input point cloud represent a certain object and the goal is to assign each point to a certain parts as shown in figure ~\ref{partseg}, hence the name "part" segmentation. In ~\cite{pointnet, sonet, DGCNN} the global descreptor learned is concateneated with the features of the points and then passed through MLP to classify each point into a part category. ~\cite{pointnetpp, pointcnn} propagates the global descreptor into high resolution predictions using interpolation and deconvolution methods respectively. In ~\cite{pointConv} the per point features learned are used to achieve segmentation by passing them through dense convolutional layers. Encoder-decoder architecture is used in ~\cite{DeepKdNet} for both parts and semantic segmenatation. In table \ref{partseg} the result of various techniques on ShapeNet parts datasets are shown.

In semantic segmentation, the goal is to assign each point to a particular class. For example, in figure ~\ref{insseg}, the points belonging to chair are shown in red, while ceiling and floor in green and blue respectively, e.t.c. Popular public datasets for Semantic segmentation evaluation are S3DIS~\cite{s3dis} and ScanNet ~\cite{dai2017scannet}. We show in table ~\ref{semanticseg} performances of some of the state-of-the-arts methods on S3DIS and ScanNet datasets.

Instance segmentation on point cloud recieves less attention compared to part and semantic segmentation. Instance segmentation is when the grouping is based on instances where multiple objects of the same class are uniquely identified. Some state-of-the-art works on instance segmentation on point cloud are ~\cite{spgn, seginstansemantics, bonet, JSIS3D, rpointnet} which are built on PointNet/PointNet++ feature learning backbone.

Table ~\ref{partseg} shows performances of the methods discussed ~\ref{nongrid} on the popular ShapeNet datasets.
\begin{table}
\begin{center}
\renewcommand{\arraystretch}{1.25}
\begin{tabular}[width=.3]{p{3cm}|p{2cm}}
\hline

\textbf{Method}&\textbf{Score} \\
\hline\hline

PointNet~\cite{pointnet} & 83.7\%\\

\hline
PointCNN~\cite{pointcnn} & 84.6\%\\
\hline
So-Net~\cite{sonet} & 84.6\%\\
\hline
PointConv~\cite{pointconvcvpr} &85.7\%\\
\hline
Kd-Network~\cite{DeepKdNet} &82.3\%\\
\hline
DGCNN~\cite{ DGCNN} &85.2\%\\
\hline
LocalSpec~\cite{localspecGCNN} &85.4\%\\
\hline
SpiderCNN~\cite{spiderCNN} & 85.3\%\\
\hline
R-S CNN~\cite{relationshape} & 86.1\%\\

\hline
A-CNN~\cite{Acnn} &86.1\%\\
\hline
ShellNet~\cite{shellnet} &82.8\%\\
\hline
InterpCNN~\cite{interpcnn} &84.0\%\\
\hline
DensePoint~\cite{densepoint} &84.2\%\\
\hline
\end{tabular}
\end{center}

\caption{Part segmentation on ShapeNet part dataset. The score is the mean Intersection Over Union(mIOU)}
\label{partseg}
\end{table}

\begin{table*}[h]
\begin{center}
\renewcommand{\arraystretch}{1.25}
\begin{tabular}[width=.5]{p{5cm}|p{3cm}|p{2cm}|p{2cm}}
\hline
\textbf{Method} &\textbf{ Datasets} & \textbf{Measure} & \textbf{Score}\\
\hline
PointNet~\cite{pointnet}& \multirow{13}{*}{S3DIS} & \multirow{8}{*}{mIOU} & 47.71\%\\
\cline{1-1} \cline{4-4}
Pointwise Conv~\cite{pointConv} & & & 56.1\%\\
\cline{1-1} \cline{4-4}
DGCNN ~\cite {DGCNN} & & & 56.1\%\\
\cline{1-1} \cline{4-4}
PointCNN~\cite{pointcnn} & & & 65.39\%\\
\cline{1-1} \cline{4-4}
PAT~\cite{selfatGSS}  & & & 54.28\%\\
\cline{1-1} \cline{4-4}
ShellNet~\cite{shellnet} & & & 66.8\%\\
\cline{1-1} \cline{4-4}
Point2Node~\cite{Point2Node} & & & 70.0\%\\
\cline{1-1} \cline{4-4}
InterpCNN~\cite{interpcnn} & & & 66.7\%\\

\cline{1-1} \cline{4-4} \cline{3-3}
PointNet~\cite{pointnet} & & \multirow{5}{*}{OA}& 78.5\%\\
\cline{1-1} \cline{4-4}
PointCNN~\cite{pointcnn} & & & 88.1\%\\
\cline{1-1} \cline{4-4}
DGCNN ~\cite {DGCNN}& & & 84.1\%\\
\cline{1-1} \cline{4-4}
A-CNN~\cite{Acnn} &&& 87.3\%\\
\cline{1-1} \cline{4-4}
JSIS3D~\cite{JSIS3D} &&& 87.4\%\\
\hline
PointNet++~\cite{pointnetpp}& \multirow{11}{*}{ScanNet} & \multirow{4}{*}{mIOU} & 55.7\%\\
\cline{1-1} \cline{4-4}
PointNet~\cite{pointnet} & & & 33.9\%\\
\cline{1-1} \cline{4-4}
PointConv~\cite{pointconvcvpr} & & & 55.6\%\\
\cline{1-1} \cline{4-4}
PointCNN~\cite{pointcnn} &&& 45.8\%\\
\cline{1-1} \cline{4-4} \cline{3-3}
PointNet~\cite{pointnet} &&\multirow{7}{*}{OA} &73.9\%\\
\cline{1-1} \cline{4-4}
PointCNN~\cite{pointcnn}&&&85.1\%\\
\cline{1-1} \cline{4-4}
A-CNN~\cite{Acnn} & & & 85.4\%\\
\cline{1-1} \cline{4-4}
LocalSpec~\cite{localspecGCNN} &&& 85.4\%\\
\cline{1-1} \cline{4-4}
PointNet++~\cite{pointnetpp}&&&84.5\%\\

\cline{1-1} \cline{4-4}
ShellNet~\cite{shellnet}&&&85.2\%\\
\cline{1-1} \cline{4-4}
Point2Node~\cite{Point2Node}&&&86.3\%\\
\hline
\end{tabular}
\end{center}
\caption{Semantic segmentation on S3DIS and ScanNet datasets. mIOU stands for mean Intersection Over Union and OA stands for Overall Accuracy}
\label{semanticseg}
\end{table*}

\subsection{Object detection}
Object detection is an extension of classification where multiple objects can be recognized and each object is localized using a bounding box as shown in figure ~\ref{objdetectionl}. RCNN ~\cite{rcnn} were the first that proposed 2D object detection by selective search, where different regions are selected and passed to the network one at a time. Several variants were later proposed ~\cite{fastrcnn, fasterrcnn, maskrcnn}. Other state-of-the-art 2D object detection is YOLO ~\cite{yolo} and its variants such as ~\cite{yolov2, yolov3}. In summary, 2D object detection is based on 2 major stages, region proposals and classification.

Like in 2D images, detection in 3D point cloud is also emperical on the two stages of proposal and classification. Proposal stage in 3D point cloud, however, is more challenging than in 2D due to the search space being 3 dimensional and the sliding window or region to be proposed is also in 3 dimension. vote3D ~\cite{vote3d} and vote3Deep ~\cite{vote3deep} convert input point cloud into a structured grid and perform extensive sliding window operation for detection which is computationally expensive. To perform object detection directly in point cloud, several techniques used feature learning techniques discussed in section ~\ref{nongrid}. 

In VoxelNet ~\cite{voxelnet}, the sparse 4D feature vectore is passed through a region proposal network to generate 3D detection. FrustumNet ~\cite{frustumnet} proposed regions in 2D and obtain the 3D frustrum of the region from the point cloud and pass it through PointNet to predict 3D bouding box. ~\cite{spgn} first uses PointNet/PointNet++ to obtain feature vector of each point, and based on the hypothesis that points belonging to the same object are closer in feature space proposed a similarity matrix which predicts if a given pair of points belong to the same object. In ~\cite{gspn}, PointNet and PointNet++ are used to designed a generative shape proposal network to generate proposals which are further processed using PointNet for classification and segmentation. PointNet++ is used in ~\cite{pointrcnn} to learn point-wise features which are used to segment foreground points from backgroud points and employs buttom-up 3D proposal to generate 3D box proposals from the foreground points. The 3D box proposals are further refined using another PointNet++-like structure. ~\cite{votenet} used PointNet++ to learn point wise features which are considered to be seeds. The seeds then independently cast a vote using a hough voting module based on MLP. The votes of the same object are close in space hence allow for easy clustering. The clusters are further processed using a shared PointNet-like module for vote aggregation and propsal. PointNet is also utilized in ~\cite{pointpillers} with Single Short Detector (SSD) ~\cite{ssd} for object detection.

One of the popular object detection dataset is the Kitti dataset ~\cite{Geiger2012CVPR, Geiger2013IJRR}. The evaluation on kitti is divided into easy, moderate and hard depending on occlusion level, minimum height of the bounding box and maximum truncation. We report the performance of various object detection methods on Kitti dataset in tables ~\ref{kitti1} and ~\ref{kitti2}.

\begin{sidewaystable*}

%\begin{table*}
  \centering

    \begin{tabular}{|c|c|c|c|c|c|c|c|c|c|c|c|c|}
    \toprule
    \multirow{2}[4]{*}{Method} & \multirow{2}[4]{*}{Modality} & \multirow{2}[4]{*}{Speed(HZ)} & mAP   & \multicolumn{3}{c|}{Car} & \multicolumn{3}{c|}{Pedestrian} & \multicolumn{3}{c|}{Cyclist} \\
\cmidrule{4-13}          &       &       & Moderate & \multicolumn{1}{l|}{Easy} & \multicolumn{1}{l|}{Moderate} & \multicolumn{1}{l|}{Hard} & \multicolumn{1}{l|}{Easy} & \multicolumn{1}{l|}{Moderate} & \multicolumn{1}{l|}{Hard} & \multicolumn{1}{l|}{Easy} & \multicolumn{1}{l|}{Moderate} & \multicolumn{1}{l|}{Hard} \\
    \midrule
    MV3D ~\cite{ChenMWLX17}  & LiDAR \& Image & 2.8   & N/A   & 86.02 & 76.9  & 68.49 & N/A   & N/A   & N/A   & N/A   & N/A   & N/A \\
    \midrule
    Cont-Fuse~\cite{LiangYWU18} & LiDAR \& Image & 16.7  & N/A   & 88.81 & 85.83 & 77.33 & N/A   & N/A   & N/A   & N/A   & N/A   & N/A \\
    \midrule
    Roarnet~\cite{ShinKT19} & LiDAR \& Image & 10    & N/A   & 88.2  & 79.41 & 70.02 & N/A   & N/A   & N/A   & N/A   & N/A   & N/A \\
    \midrule
    AVOD-FPN~\cite{KuMLHW18} & LiDAR \& Image & 10    & 64.11 & 88.53 & 83.79 & 77.9  & 58.75 & 51.05 & 47.54 & 68.09 & 57.48 & 50.77 \\
    \midrule
    F-PointNet~\cite{frustumnet} & LiDAR \& Image & 5.9   & 65.39 & 88.7  & 84    & 75.33 & 58.09 & 50.22 & 47.2  & 75.38 & 61.96 & 54.68 \\
    \midrule
    HDNET~\cite{YangLU18} & LiDAR \& Map & 20    & N/A   & 89.14 & 86.57 & 78.32 & N/A   & N/A   & N/A   & N/A   & N/A   & N/A \\
    \midrule
    PIXOR++~\cite{YangLUcvpr18} & LiDAR & 35    & N/A   & 89.38 & 83.7  & 77.97 & N/A   & N/A   & N/A   & N/A   & N/A   & N/A \\
    \midrule
    VoxelNet~\cite{voxelnet} & LiDAR & 4.4   & 58.52 & 89.35 & 79.26 & 77.39 & 46.13 & 40.74 & 38.11 & 66.7  & 54.76 & 50.55 \\
    \midrule
    SECOND~\cite{YanMLsensors18} & LiDAR & 20    & 60.56 & 88.07 & 79.37 & 77.95 & 55.1  & 46.27 & 44.76 & 73.67 & 56.04 & 48.78 \\
    \midrule
    PointRCNN~\cite{ShiWL19} & LiDAR & N/A   & N/A   & 89.28 & 86.04 & 79.02 & N/A   & N/A   & N/A   & N/A   & N/A   & N/A \\
    \midrule
    PointPillars~\cite{LangVCZYB19} & LiDAR & 62    & 66.19 & 88.35 & 86.1  & 79.83 & 58.66 & 50.23 & 47.19 & 79.14 & 62.25 & 56 \\
    \bottomrule
    \end{tabular}%
  \caption{Performance on the KITTI Bird’s Eye View detection benchmark}
  \label{kitti1}%
%\end{table*}%
\vspace{2cm}

%\begin{table*}
  \centering

    \begin{tabular}{|c|c|c|c|c|c|c|c|c|c|c|c|c|}
    \toprule
    \multirow{2}[4]{*}{Method} & \multirow{2}[4]{*}{Modality} & \multirow{2}[4]{*}{Speed(HZ)} & mAP   & \multicolumn{3}{c|}{Car} & \multicolumn{3}{c|}{Pedestrian} & \multicolumn{3}{c|}{Cyclist} \\
\cmidrule{4-13}          &       &       & Moderate & \multicolumn{1}{l|}{Easy} & \multicolumn{1}{l|}{Moderate} & \multicolumn{1}{l|}{Hard} & \multicolumn{1}{l|}{Easy} & \multicolumn{1}{l|}{Moderate} & \multicolumn{1}{l|}{Hard} & \multicolumn{1}{l|}{Easy} & \multicolumn{1}{l|}{Moderate} & \multicolumn{1}{l|}{Hard} \\
    \midrule
    MV3D  ~\cite{ChenMWLX17}  & LiDAR \& Image & 2.8   & N/A   & 71.09 & 62.35 & 55.12 & N/A   & N/A   & N/A   & N/A   & N/A   & N/A \\
    \midrule
    Cont-Fuse ~\cite{LiangYWU18} & LiDAR \& Image & 16.7  & N/A   & 82.54 & 66.22 & 64.04 & N/A   & N/A   & N/A   & N/A   & N/A   & N/A \\
    \midrule
    Roarnet~\cite{ShinKT19} & LiDAR \& Image & 10    & N/A   & 83.71 & 73.04 & 59.16 & N/A   & N/A   & N/A   & N/A   & N/A   & N/A \\
    \midrule
    AVOD-FPN~\cite{KuMLHW18} & LiDAR \& Image & 10    & 55.62 & 81.94 & 71.88 & 66.38 & 50.8  & 42.81 & 40.88 & 64    & 52.18 & 46.64 \\
    \midrule
    F-PointNet~\cite{frustumnet} & LiDAR \& Image & 5.9   & 57.35 & 81.2  & 70.39 & 62.19 & 51.21 & 44.89 & 40.23 & 71.96 & 56.77 & 50.39 \\
    \midrule
    VoxelNet~\cite{voxelnet} & LiDAR & 4.4   & 49.05 & 77.47 & 65.11 & 57.73 & 39.48 & 33.69 & 31.5  & 61.22 & 48.36 & 44.37 \\
    \midrule
    SECOND~\cite{YanMLsensors18} & LiDAR & 20    & 56.69 & 83.13 & 73.66 & 66.2  & 51.07 & 42.56 & 37.29 & 70.51 & 53.85 & 46.9 \\
    \midrule
    PointRCNN~\cite{ShiWL19} & LiDAR & N/A   & N/A   & 84.32 & 75.42 & 67.86 & N/A   & N/A   & N/A   & N/A   & N/A   & N/A \\
    \midrule
    PointPillars~\cite{LangVCZYB19} & LiDAR & 62    & 59.2  & 79.05 & 74.99 & 68.3  & 52.08 & 43.53 & 41.49 & 75.78 & 59.07 & 52.92 \\
    \bottomrule
    \end{tabular}%
  \caption{Performance on the KITTI 3D object detection benchmark}
  \label{kitti2}%
%\end{table*}%

\end{sidewaystable*}

\section{Summary and Conclusion}
The increasing availability of point cloud as a result of evolving scanning devices coupled with increasing application in autonomous vehicles, robotics, AR and VR demands for fast and efficient algorithms for the point cloud processing inorder to achieve improved visual perception such as recognition, segmentation and detection. Due to scarse data availability, unpopularity of deep learning, early methods for point cloud processing relied on handcrafted features. However, with the revolution brought about by deep learning in 2D vision tasks, and evolution of acquisition devices of point cloud which leads to availability of point cloud data, computer vision community are focusing on how to utilize the power of deep learning on point cloud data. Point cloud provides more accurate 3D information which is vital in applications that require 3D information. Due to the nature of point cloud, its very challenging to use deep learning for its processing. Most approaches resolve to convert the point cloud into a structured grid for easy processing by deep neural networks. These approaches, however, leads to either loss of depth information or  introduces conversion artifacts and requires higher computational cost. Recently, deep learning directly on point cloud is recieving alot of attention. Learning on point cloud directly do away with convertion artifacts and mitigates the need for higher computational cost. PointNet is the basic deep learning method that process point cloud directly. PointNet however, does not capture local structures. Many approaches were developed to improve on pointNet by capturing local structures. Inorder to capture local structures, most methods follows three basic steps; sampling to reduce the resolution of points and to get centroids for representing local neighborhood; grouping, based on K-NN to select neighboring points to each centroids; mapping function, usually approximated by an MLP, which learn the representation of neigbhoring points. Several methods resolves to approximating the MLP with PointNet-like network. However because PointNet does not explore inter points relationship, several approaches explore inter-points relationships within a local patch before applying pointNet like MLPs. Taking into account the point-to-point relationship between points has proven to increase the discriminative capability of the networks.

While deep learning on 3D point cloud has shown good performance on several tasks including classification, parts and semantic segmentation, other areas, however, are recieving less attention. Instance segmentation on 3D point cloud, where individual objects are segmented in a scene, remain largely uncharted direction. Most current object detection relies on 2D detection for region proposal, few works are available on detecting objects directly on point cloud. Scaling to larger scene also remain largely unexploited as most of the current works relies on cutting large scenes into smaller pieces. As at the time of this review, only few works ~\cite{pointnetvlad, lpdnet} explored deep learning on large scale 3D scene.

\bibliographystyle{elsarticle-num-names}

\bibliography{mybibfile}

\end{document}